\documentclass[10pt,twocolumn,letterpaper]{article}

\usepackage[pagenumbers]{cvpr}

\definecolor{cvprblue}{rgb}{0.21,0.49,0.74}
\usepackage[pagebackref,breaklinks,colorlinks,allcolors=cvprblue]{hyperref}

\title{TopoPerception: A Shortcut-Free Evaluation of Global Visual Perception in Large Vision-Language Models}

\author{
Wenhao Zhou \quad Hao Zheng \quad Rong Zhao\thanks{Corresponding author.}\\
Center for Brain-Inspired Computing Research (CBICR), Tsinghua University, Beijing, China\\
Department of Precision Instruments, Tsinghua University, Beijing, China\\
IDG/McGovern Institute for Brain Research, Tsinghua University, Beijing, China\\
{\tt\small zwh20@mails.tsinghua.edu.cn, hao\_z@mit.edu, r\_zhao@mail.tsinghua.edu.cn}
}

\begin{document}
\maketitle
\begin{abstract}
Large Vision-Language Models (LVLMs) typically align visual features from an encoder with a pre-trained Large Language Model (LLM). However, this makes the visual perception module a bottleneck, which constrains the overall capabilities of LVLMs. Conventional evaluation benchmarks, while rich in visual semantics, often contain unavoidable local shortcuts that can lead to an overestimation of models' perceptual abilities. Here, we introduce TopoPerception, a benchmark that leverages topological properties to rigorously evaluate the global visual perception capabilities of LVLMs across various granularities. Since topology depends on the global structure of an image and is invariant to local features, TopoPerception enables a shortcut-free assessment of global perception, fundamentally distinguishing it from semantically rich tasks. We evaluate state-of-the-art models on TopoPerception and find that even at the coarsest perceptual granularity, all models perform no better than random chance, indicating a profound inability to perceive global visual features. Notably, a consistent trend emerge within model families: more powerful models with stronger reasoning capabilities exhibit lower accuracy. This suggests that merely scaling up models is insufficient to address this deficit and may even exacerbate it. Progress may require new training paradigms or architectures. TopoPerception not only exposes a critical bottleneck in current LVLMs but also offers a lens and direction for improving their global visual perception. The data and code are publicly available at: \url{https://github.com/Wenhao-Zhou/TopoPerception}.
\end{abstract}    
\section{Introduction}
\label{sec:intro}

The field of artificial intelligence (AI) has undergone a revolutionary transformation with the advent of Large Vision-Language Models (LVLMs). Recent state-of-the-art (SOTA) models combine advanced visual understanding with the powerful reasoning capabilities of Large Language Models (LLMs), demonstrating remarkable performance on a wide range of multimodal tasks, including Visual Question Answering (VQA), captioning, and complex visual reasoning~\cite{openai2024gpt4technicalreport, gpt4vsystemcard, openai2024gpt4ocard, o3o4systemcard, Claude3systemcard, Claude37systemcard, Claude4systemcard, geminiteam2025geminifamilyhighlycapable, geminiteam2024gemini15unlockingmultimodal, comanici2025gemini25pushingfrontier}. The dominant architecture in these models involves a powerful visual encoder that processes visual inputs into a set of feature vectors or tokens, which are then fed through a small projection or query network into an LLM to generate responses based on the visual content~\cite{10.1093/nsr/nwae403, li2025surveystateartlarge}.

\begin{figure}[t]
\centering
\includegraphics[width=1\columnwidth]{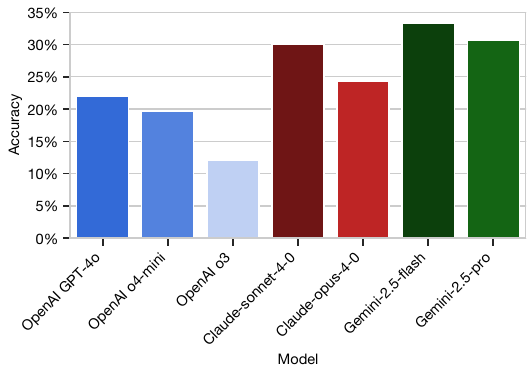}
\caption{Accuracy of various LVLMs on the easiest level of TopoPerception. Models from the same family are represented by the same color scheme.}
\label{fig1}
\end{figure}

\begin{figure*}[t]
\centering
\includegraphics[width=1\textwidth]{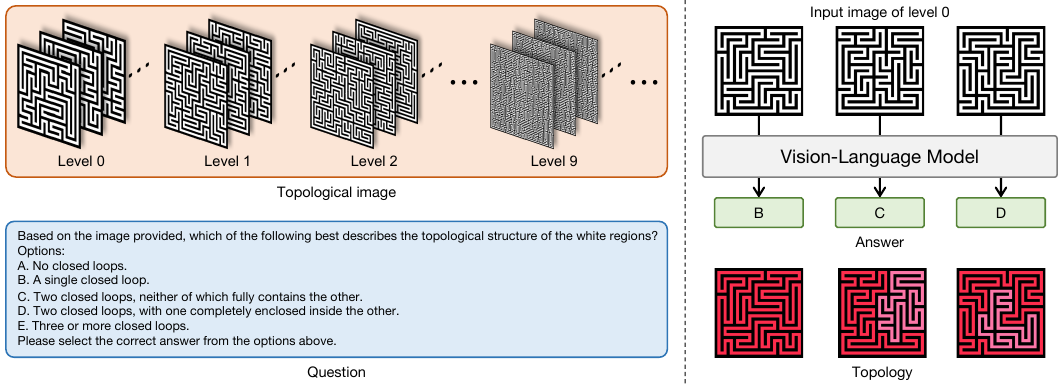}
\caption{An illustration of the TopoPerception benchmark. The text question and options are fixed (left). The input image belongs to one of three categories, corresponding to options B, C, and D, respectively (right). Options A and E serve as distractors. The input images can be extended to arbitrary difficulty levels.}
\label{fig2}
\end{figure*}

Despite its success, it introduces a critical and often overlooked perceptual bottleneck. The conversion of a high-dimensional, continuous visual data into a discrete sequence of tokens is inherently a lossy compression process. Semantically, the training objectives for many visual encoders, such as contrastive learning in models like CLIP~\cite{pmlr-v139-radford21a}, aim to create representations that align with natural language descriptions. However, this semantic compression process can inadvertently discard crucial global visual information that lacks a straightforward linguistic correlate. Structurally, the design of visual encoders imposes strong inductive biases, such as fixed input resolutions and aspect ratios, which often necessitate scaling or patching the image~\cite{NEURIPS2024_5e221748}. Such operations can corrupt global structures and distort spatial relationships, compromising the integrity of the visual representation before it even reaches the language model. Furthermore, long sequences of visual tokens impose a significant computational burden, motivating research into token reduction techniques~\cite{shao2025tokenstalkmuchsurvey} that may further risk discarding subtle yet important visual details.

Most current evaluation methods for LVLMs, while comprehensive, are predominantly focused on downstream, semantically rich tasks. Benchmarks for VQA, compositional reasoning, and open-world dialogue assess a combination of perception, reasoning, and language generation capabilities~\cite{li2024surveybenchmarksmultimodallarge}. However, models' success or failure on these tasks does not isolate the fidelity of its initial visual perception. This capability confounding can mask fundamental weaknesses. For instance, models might answer questions by leveraging parameterized knowledge stored within the LLM rather than relying on the visual input. This stored knowledge may be correct~\cite{balanced_vqa_v2, 8099698, 8953451, NEURIPS2024_2f8ee6a3} or incorrect~\cite{li-etal-2023-evaluating, 10657594}. Even the design of the question itself may be flawed, allowing a correct answer to be derived from the input text alone, without any need for the visual input~\cite{liu2024mmbench, NEURIPS2024_2f8ee6a3}.

Moreover, these semantically rich tasks introduce unavoidable local shortcuts, where models might correctly answer questions by identifying a specific object or scene element without understanding the image's global structure~\cite{10.1371/journal.pcbi.1006613,geirhos2018imagenettrained, Geirhos2020, NEURIPS2024_a13ff984, gavrikov2025can}. This reliance on local shortcuts can lead to an overestimation of the model's true visual perception, as it may be exploiting local cues instead of processing the image globally. Therefore, there is an urgent need for an evaluation paradigm that can disentangle visual perception from textual reasoning and directly measure global visual perception in a shortcut-free manner.

Topology offers a promising avenue for addressing this challenge. As a branch of mathematics, topology studies the properties of space that are preserved under continuous deformations, such as stretching, twisting, crumpling, and bending~\cite{Munkres2000Topology}. When considering the two-dimensional manifold projected onto the human retina, these invariant properties can be categorized into three main topological attributes: connectivity, the number of holes, and interior/exterior relationships. Crucially, these attributes are independent of local features. Therefore, topological features capture the global structure of an image, making them an ideal proxy for evaluating models' global visual perception without the influence of local shortcuts. Furthermore, topology is closely related to processes in the human visual system. For humans, global features enable rapid object detection without requiring an in-depth analysis of local features, a critical skill for survival in natural environments. Given the pronounced sensitivity of the human visual system to global features, it has been hypothesized that global features are processed first and subsequently guide the interpretation of local features~\cite{Chen1982, Chen2005}.

In this paper, we introduce TopoPerception, a diagnostic benchmark that employs the topological features of images to isolate and quantify global visual perception without shortcuts. To isolate the visual input and prevent the model from relying on textual cues to answer, TopoPerception uses a fixed text-based question and a fixed set of options that do not change with the visual input. Furthermore, to ensure that no local visual shortcuts are introduced, TopoPerception utilizes synthetically generated images that possess pure topological properties without confounding semantic features. As previously stated, because topological properties are independent of local features, TopoPerception can precisely control the granularity of the global features being evaluated without being susceptible to local shortcuts. By testing whether LVLMs can differentiate between images based on their topological class across various levels of perceptual granularity, TopoPerception probes whether the models' visual encoder and subsequent alignment process preserve the global features of an image or merely convert them into a language-compatible format at the expense of global fidelity.

The main contributions of this study are threefold:
\begin{itemize}
\item We present TopoPerception, a shortcut-free benchmark for evaluating the global visual perception of LVLMs. It defines tasks based on topological properties across a range of perceptual granularities, providing a scalable difficulty hierarchy for stress-testing LVLMs.
\item We conduct a comprehensive evaluation of SOTA LVLMs, revealing systematic and severe deficiencies in their global visual perception. Even at the simplest, coarsest evaluation granularity in TopoPerception, all models performed at a level statistically consistent with random guessing, answering almost entirely based on their intrinsic biases (\cref{fig1}). This highlights a substantial loss of global visual features in current LVLMs.
\item We discover an unexpected trend: within the same model family or architecture, larger models with stronger reasoning capabilities tend to exhibit lower accuracy on TopoPerception (\cref{fig1}). In other words, prompting models to generate more detailed, step-by-step answers is positively correlated with a decrease in accuracy when perceiving global visual features. This suggests that scaling up reasoning may actually interfere with or override the models' already fragile visual signal, a finding that raises concerns about the interplay between chain-of-thought reasoning and visual grounding.
\end{itemize}

\section{Related work}
\label{relate}

\subsection{Evaluation benchmarks for LVLMs}
The rapid development of LVLMs has been accompanied by the establishment of a diverse ecosystem of evaluation benchmarks, each designed to probe different dimensions of multimodal intelligence~\cite{liu2024mmbench, fu2024mmecomprehensiveevaluationbenchmark, MMMU, 10.5555/3692070.3694451}. While this rich ecosystem has been instrumental in driving progress, these benchmarks share a common orientation: they predominantly evaluate the semantic interpretation of visual content. Success in these tasks is contingent on the correctness of a textual response to a query about the meaning, composition, or context of an image. However, the complexity of semantic content introduces unavoidable local shortcuts. This leaves a critical gap in our understanding of a more fundamental, upstream capability: the fidelity of global visual perception itself. TopoPerception aims to fill this gap by directly measuring the degree to which global visual features are preserved, independent of their semantic content.

\begin{figure}[t]
\centering
\includegraphics[width=1\columnwidth]{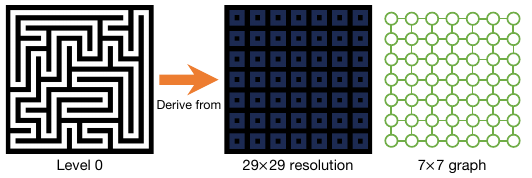}
\caption{The visual topological images are generated by constructing a uniform spanning tree on a connected graph. Each node in the graph corresponds to a $3\times 3$ pixel block in the image. A connected graph with $n\times n$ nodes generates an image with a resolution of $(4n+1)\times (4n+1)$.}
\label{fig3}
\end{figure}

\begin{figure*}[t]
\centering
\includegraphics[width=1\textwidth]{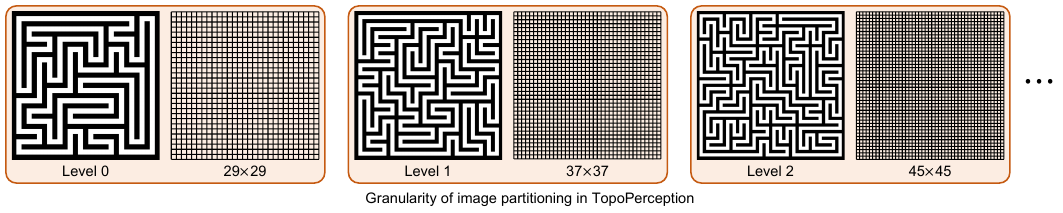}
\caption{Illustration of the granularity of image partitions at different levels in TopoPerception. The difficulty level for an image with a partitioning granularity of $(4n+1)\times (4n+1)$ is defined as $(n-7)/2$.}
\label{fig4}
\end{figure*}

\subsection{Benchmarking global perception in LVLMs}
Current benchmarks for evaluating the global perception capabilities of LVLMs can generally be classified into two main categories:
\begin{itemize}
    \item \textbf{Shape-based}: These methods indirectly express global attributes through the shape of objects~\cite{gavrikov2025can,NEURIPS2024_a13ff984}. However, the shape itself may contain local shortcuts, such as distinguishing between circles and squares, which can be determined based on the local curvature of the shape rather than its overall form.
    \item \textbf{Maze-based}: Maze tasks typically specify a start and end point and assess the connectivity between them~\cite{lotfi2025chainofsketchenablingglobalvisual}. When the points are not connected, the model must explore all possible paths to determine the result. In contrast, when the points are connected, the model only needs to identify one valid solution, which significantly reduces the demand for global perception. This introduces local shortcuts in the evaluation of global perception abilities and makes it difficult to control the difficulty of the dataset.
\end{itemize}
In contrast, TopoPerception utilizes topological properties, which effectively eliminate the possibility of local shortcuts, offering a more robust evaluation of global perception. This innovation allows us to directly assess the model's ability to preserve and process the global structure of an image, bypassing the confounding influence of local cues.

\subsection{The visual pipeline in LVLMs: sources of information loss}
The standard architecture of an LVLM consists of three main components: a visual encoder, a language model, and a projection module to connect the two~\cite{NEURIPS2022_960a172b, NEURIPS2023_e425b75b, peng2024grounding, NEURIPS2023_6dcf277e, Liu_2024_CVPR, li2025llavaonevision, bai2023qwenvlversatilevisionlanguagemodel, wang2024qwen2vlenhancingvisionlanguagemodels, bai2025qwen25vltechnicalreport, xu2025qwen25omnitechnicalreport, lu2024deepseek, wu2024deepseek, grattafiori2024llama3herdmodels, guo2025seed15vltechnicalreport, vteam2025glm41vthinkingversatilemultimodalreasoning, Wu_2025_CVPR, chen2025janusprounifiedmultimodalunderstanding}. The visual encoder, responsible for converting visual inputs into a sequence of feature vectors or tokens, is a critical source where significant information loss can occur. Pre-trained visual encoders often introduce strong inductive biases, such as being trained on fixed-size square images. To handle arbitrarily shaped inputs, models must resort to operations like resizing, padding, or patching the image~\cite{NEURIPS2024_5e221748}, which can severely distort the image's global layout and corrupt the structural relationships between different parts of a scene, leading to a compromised representation before it is processed by the language model. Furthermore, as the image is transformed into a dense sequence of tokens, the model may risk discarding important visual details in an effort to reduce the computational load~\cite{shao2025tokenstalkmuchsurvey}, compounding the challenge of accurately preserving the global visual structure.

Subsequently, the long sequence of visual tokens is passed through the projection module and into the LLM. A significant challenge at this stage is the cross-modal alignment problem, where the learned visual representations may not effectively correspond to textual concepts in the LLM's embedding space. This alignment discrepancy can complicate information fusion and may undermine the effectiveness of techniques aimed at reducing the computational cost of long visual sequences, such as token reduction~\cite{xu2025rethinkingvisualtokenreduction}. Our research provides a new diagnostic tool to precisely quantify a key consequence of these architectural choices: the loss of globally perceived visual features.

\section{Sources of shortcuts in benchmarks}
\label{source}

We classify the potential shortcuts in LVLM evaluation benchmarks into two hierarchical levels:
\begin{itemize}
\item \textbf{Statistical level}: Due to improper data distribution design, the data may contain unintended statistical regularities, i.e., spurious correlations. This is commonly observed in training-dependent scenarios~\cite{Geirhos2020, Zhou2025}. For example, in vision-language benchmarks, visual elements frequently co-occur with corresponding textual cues, allowing the task to be solved correctly by relying solely on the language modality without genuinely understanding the image content~\cite{balanced_vqa_v2}. In this case, the language modality itself becomes a shortcut. Another example is when certain elements in an image frequently appear together; the model can infer the presence of one element from the other. Here, the co-occurring elements are the shortcut~\cite{li-etal-2023-evaluating}.
\item \textbf{Semantic level}: This level of shortcuts does not depend on statistical co-occurrence but can be inferred from the semantics of the inputs alone. For instance, in a VQA task, the text of the question might contain sufficient information to be answered without reference to the image content~\cite{liu2024mmbench, NEURIPS2024_2f8ee6a3}. Here, the textual content itself acts as the shortcut. As another example, in an LVLM recognition task, we could ask in two ways:
\begin{enumerate}
    \item ``What is the category of this image?''
    \item ``Based on its texture, what is the category of this image?''
\end{enumerate}
Compared to the second question, the first allows the model to rely on shape instead of texture, making shape a potential shortcut. The second question explicitly eliminates this shortcut at the prompt level~\cite{gavrikov2025can}. It is clear that semantic-level shortcuts can be identified from a single sample without resorting to statistical patterns. However, they also manifest as statistical regularities, meaning the semantic level is contained within the statistical level. Therefore, semantic shortcuts exist in both training-dependent and training-free scenarios~\cite{Zhou2025}. Since LVLM benchmarks are typically zero-shot, they are primarily susceptible to semantic-level shortcuts.
\end{itemize}

\section{TopoPerception benchmark}
\label{benchmark}

\subsection{Benchmark design and task format}
As discussed, LVLM benchmarks often face semantic-level shortcuts. To accurately evaluate the global visual perception of LVLMs without being affected by local shortcuts at a semantic level, TopoPerception constructs its tasks using topological properties of images, which are by definition independent of local features. Furthermore, we have designed the benchmark to prevent models from using parameterized knowledge stored from prior training data by addressing both the visual and textual modalities:
\begin{itemize}
\item \textbf{Visual modality}: To prevent models from memorizing relevant visual features of natural images, we use a synthetic dataset as the foundation for our visual data.
\item \textbf{Textual modality}: To isolate the visual input and ensure that the model cannot answer based solely on the text, we employ a fixed text question and a fixed set of answer options that remain constant across all visual inputs.
\end{itemize}

\Cref{fig2} (left) illustrates the task format of TopoPerception, which takes the form of a multiple-choice VQA task. Each query presented to the model consists of one image and a fixed text question with five options. The question asks about the topological properties of the image and provides five lettered options (A, B, C, D, E) as possible answers. The visual modality uses a synthetic dataset of topological images~\cite{https://doi.org/10.6084/m9.figshare.28794407}, which not only prevents models from relying on memorized natural image features but also offers highly scalable difficulty levels. According to Kirchhoff's theorem~\cite{https://doi.org/10.1002/andp.18471481202}, the topological image sample space exhibits asymptotic exponential growth $\sim \exp(h n^2)$ with respect to the number of nodes $n$, where $h \approx 1.16624$ is the lattice tree entropy constant~\cite{Robert_Shrock_2000}. This ensures that even if TopoPerception is later included in training data, the model cannot solve its extended difficulty levels through memorization. The images have three categories of topological properties, as shown in \cref{fig2} (right). The textual modality is a fixed multiple-choice question asking about the image's topological properties, ensuring that the only variable is the image itself. Therefore, the model must rely on the image to select the correct option.

\begin{figure}[t]
\centering
\includegraphics[width=1\columnwidth]{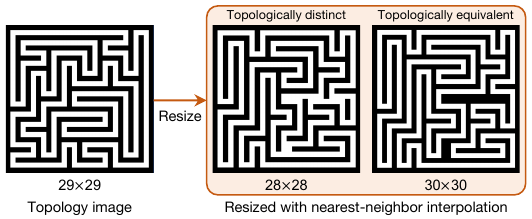}
\caption{An example of how the topological properties of an image in TopoPerception change across adjacent perceptual granularities.}
\label{fig5}
\end{figure}

Options B, C, and D in the question correspond to the three topological categories of the images. Options A and E serve as distractors to ensure the question is closed-form, meaning all possible correct answers are among the options. This helps diagnose whether the model is guessing randomly or exhibits a bias towards certain options. If a model's accuracy is close to 20\%, it suggests random guessing among all five options (A, B, C, D, E). If the accuracy is close to 33.3\%, it suggests a preference for certain options within the valid set (B, C, D).

\begin{table*}[t]
\centering
\begin{tabular}{l|ccc|c}
\toprule
Model & Accuracy (\%) & Precision (\%) & Recall (\%) & F1 Score (\%) \\
\midrule
OpenAI GPT-4o~\cite{openai2024gpt4ocard} & 22.00 & 33.40 & 22.00 & 24.07 \\
OpenAI o4-mini~\cite{o3o4systemcard} & 19.67 & 28.28 & 19.67 & 21.59 \\
OpenAI o3~\cite{o3o4systemcard} & 12.00 & 35.36 & 12.00 & 16.38 \\
\midrule
Claude-sonnet-4-0~\cite{Claude4systemcard} & 30.00 & 42.51 & 30.00 & \textbf{28.07} \\
Claude-opus-4-0~\cite{Claude4systemcard} & 24.33 & 21.50 & 24.33 & 22.06 \\
\midrule
Gemini-2.5-flash~\cite{comanici2025gemini25pushingfrontier} & \textbf{33.33} & \textbf{48.44} & \textbf{33.33} & 26.34 \\
Gemini-2.5-pro~\cite{comanici2025gemini25pushingfrontier} & 30.67 & 33.11 & 30.67 & 27.96 \\
\bottomrule
\end{tabular}
\caption{Results of LVLMs on TopoPerception at the lowest difficulty level (Level 0). The best result in each column is bolded. Precision, Recall, and F1 Score are weighted averages of the per-class metrics, weighted by the number of true samples in each class.}
\label{table1}
\end{table*}

\subsection{Topological properties and granularity levels}
As shown in \cref{fig3}, the visual topological images used in TopoPerception are generated by constructing a uniform spanning tree on a connected graph, which includes multiple difficulty levels~\cite{Zhou2025}. Each difficulty level corresponds to a different spatial resolution, i.e., a different granularity of partitioning for the image, as illustrated in \cref{fig4}. At a lower partitioning granularity, the image is divided into coarser regions, placing weaker demands on the model's information retention capacity. At a finer granularity, the demands are stronger. By varying the granularity, TopoPerception allows us to control the difficulty of the task and assess the model's ability to retain global visual information at different levels of abstraction.

As previously mentioned, since topological properties are determined by the global structure of the image and are independent of local features, a model can only preserve the image's topological properties if its perceptual granularity is greater than or equal to the image's own partitioning granularity. To demonstrate this intuitively, we analyze how the topological properties of an image change across two adjacent perceptual granularities, as shown in \cref{fig5}. Using nearest-neighbor downsampling as an example, when the model's perceptual granularity (i.e., the resolution it effectively processes) is smaller than the image's partitioning granularity, the topological properties may be distorted or lost. Conversely, when the model's perceptual granularity is greater than or equal to the image's partitioning granularity, the topological properties remain stable. This insight allows us to precisely determine the model's true ability to retain global visual information as it processes an image at different levels of resolution during perception.

\section{Experiments}
\label{experiment}

\subsection{Experimental setup}
We selected some of the most powerful recent LVLMs for our evaluation:
\begin{itemize}
\item \textbf{OpenAI}: GPT-4o~\cite{openai2024gpt4ocard}, o4-mini, o3~\cite{o3o4systemcard}.
\item \textbf{Anthropic}: Claude-sonnet-4-0, Claude-opus-4-0~\cite{Claude4systemcard}.
\item \textbf{Google}: Gemini-2.5-flash, Gemini-2.5-pro~\cite{comanici2025gemini25pushingfrontier}.
\end{itemize}
All models were evaluated using their standard API calls. We chose these models because they represent the current SOTA and have demonstrated outstanding performance on various vision-language tasks.

For each difficulty level and each category of the visual topology dataset, we randomly sampled 100 images to construct the TopoPerception benchmark. This resulted in 300 samples per granularity level, balancing cost with statistical significance.

To observe the processes for the models' answers, we did not force them to output only the option letter but allowed them to generate open-ended text responses. We used the default inference settings for each model. It is worth noting that, since all textual questions and answer options were identical across models, we retained the default temperature rather than setting it to 0, in order to preserve the natural stochasticity of the model's generative distribution. This approach enables us to further investigate potential preference patterns when a model shows inclination towards particular options, thereby facilitating an analysis of the strength and stability of its biases across the entire output distribution. By doing so, we can uncover systematic differences at the probabilistic level, rather than being limited to deterministic outputs, thus providing a more comprehensive characterization of the model's behavioral tendencies.

\subsection{Results on the TopoPerception}
\Cref{table1} summarizes the performance of the evaluated models on the TopoPerception benchmark at Level 0, which corresponds to a resolution of $29\times 29$. This resolution is comparable to the MNIST dataset's $28\times 28$ resolution~\cite{726791}, providing a straightforward reference for difficulty. \Cref{fig1} provides an intuitive comparison of the accuracy of each model, with models from the same family indicated by a consistent color scheme. Remarkably, the performance of most models hovers around the 20\% random baseline, indicating that their accuracy is statistically indistinguishable from random guessing. No model's performance exceeds the 33.3\% accuracy threshold that would result from a perfect bias towards one of the three correct options. This highlights a significant gap in the current LVLMs' ability to comprehend the global structure of an image. Even the simplest evaluation task, reveals a profound inability to process visual information at the global level.

Notably, \Cref{fig1} reveals that within each model family, variants are close in accuracy. However, a consistent trend is discernible: larger models with more powerful reasoning capabilities tend to have lower accuracy on TopoPerception. This phenomenon holds across different model families. For instance, in the OpenAI family, the accuracy follows GPT-4o $>$ o4-mini $>$ o3. In the Anthropic family, Claude-sonnet-4-0 outperforms Claude-opus-4-0. In the Google family, Gemini-2.5-flash is more accurate than Gemini-2.5-pro. This trend suggests that while increased reasoning ability of models may boost performance on other tasks, they do not automatically improve visual perception capabilities. In fact, it may even exacerbate the issue.

We hypothesize that because topological features are global and abstract, LVLMs that attempt to reason through natural language may be misled by their own priors or by the absence of a precise internal representation of the image. Models with stronger reasoning abilities may be more reliant on language-based reasoning, which could distort their visual understanding. This counter-intuitive finding suggests that for tasks requiring strict perceptual fidelity, prompting step-by-step reasoning is not always beneficial, unlike in arithmetic or commonsense problems. Instead, a more direct vision-to-decision mapping, perhaps akin to a classification head, might be more effective.

\begin{figure*}[t]
\centering
\includegraphics[width=1\textwidth]{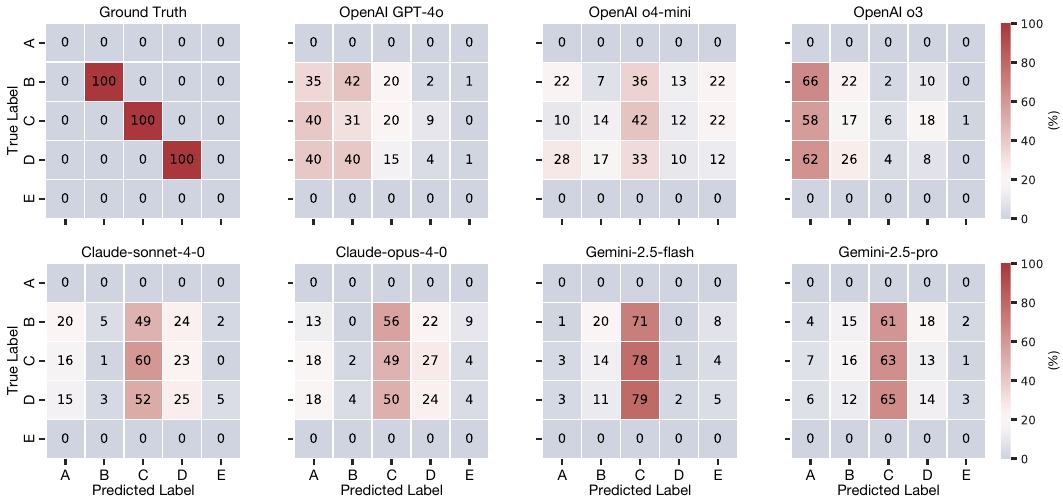}
\caption{Confusion matrices for all evaluated LVLMs on the easiest level of TopoPerception.}
\label{fig6}
\end{figure*}

\subsection{Error analysis}
To further investigate where the models are failing, we present the confusion matrices for all models on the Level 0 task, as shown in \cref{fig6}. It is evident that the models exhibit significant biases on TopoPerception, and models from the same family show similar tendencies. For example, both models in the Claude 4 family, Claude-sonnet-4-0 and Claude-opus-4-0, are most inclined to choose option C, followed by options A and D. Similarly, both models in the Gemini 2.5 family, Gemini-2.5-flash and Gemini-2.5-pro, show a strong preference for option C. In contrast, the preferences of the OpenAI family exhibit more varied behavior, likely because they belong to different series. GPT-4o primarily favors options A and B, followed by C; o4-mini's choices are more uniform, resembling random guessing, which is corroborated by its 19.67\% accuracy; o3, on the other hand, has a very strong bias towards option A.

\begin{figure*}[t]
\centering
\includegraphics[width=1\textwidth]{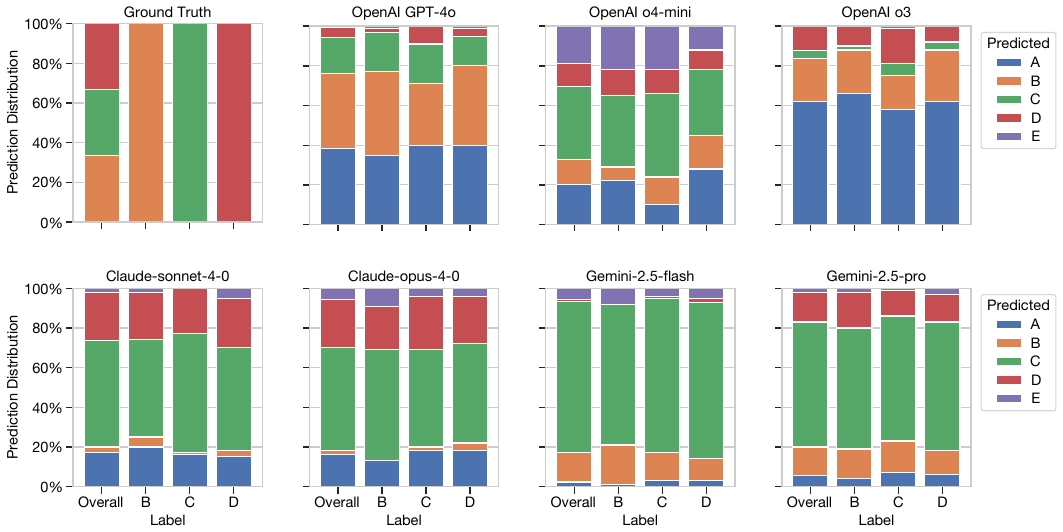}
\caption{Prediction distributions of LVLMs for data from different ground-truth labels on the easiest level of TopoPerception. ``Overall'' represents the prediction distribution across all test data.}
\label{fig7}
\end{figure*}

To gain deeper insight into the source of these biases, we analyze the prediction distributions for each model on each image category for the Level 0 task, as shown in \cref{fig7}. It can be observed that for each model, the prediction distribution is nearly identical across different categories and consistent with the overall data prediction distribution. This indicates that all models are using their inherent biases to select answers for every question, and their selection strategy does not change based on the input image. Combined with our choice of temperature parameter, we can conclude that this preference is a stable characteristic of the models.

\section{Discussion}
\label{discussion}

\subsection{Classification of shortcuts in benchmarks}
In addition to the previously described classification of shortcuts into statistical and semantic levels, we can also categorize shortcuts in LVLM evaluation benchmarks according to a ``Morgan's Canon of data''~\cite{Zhou2025}:
\begin{itemize}
    \item \textbf{Type 1}: Tasks can be solved using only the text modality, without requiring input from the visual modality. In this case, the text modality itself becomes the shortcut.
    \item \textbf{Type 2}: Tasks require visual input, but there is a shortcut within the image itself, such as a local feature shortcut.
\end{itemize}
Therefore, eliminating Type 2 shortcuts presupposes the elimination of Type 1 shortcuts. Conversely, eliminating Type 1 shortcuts does not affect the existence of Type 2 shortcuts. TopoPerception eliminates Type 1 shortcuts through its fixed question-and-option design and then eliminates Type 2 shortcuts through the introduction of topological properties.

\subsection{Synthetic data}
Some may argue that the use of synthetic data in TopoPerception, which differs significantly from the distribution of natural images. Therefore, one might question why we expect LVLMs to perform well. We emphasize that, on one hand, natural data often contains a complex mixture of properties that make it difficult to assess a specific attribute clearly. In contrast, synthetic data serves as an abstraction of specific objects or properties, designed to decouple and control these factors---something that natural images cannot achieve. Consequently, synthetic and natural data complement each other in scientific research, with synthetic data enabling the evaluation of properties that cannot be directly assessed in natural data. On the other hand, the goal of synthetic data is to minimize interference from prior learning experiences. This concept is similar to how fluid intelligence is measured in human IQ tests, such as Raven's Progressive Matrices~\cite{Cattell1943, John2003}. In these tests, researchers design abstract tasks, rather than those based on real-world scenarios, to evaluate cognitive ability. These tasks do not depend on specific knowledge or experience and are applicable across various age groups. In the context of TopoPerception, the introduction of synthetic data represents a pure abstraction of global properties, preventing the model from memorizing certain visual features in natural images that could compromise the reliability of the evaluation.

\subsection{Interpretation of experimental results}
The consolidated findings of our study paint a concerning picture: despite the large scale of current LVLMs and their training on vast image-text corpora, these models fail to preserve or reason about global visual features reliably. A critical aspect of this problem lies in the role of model scale and architecture. We found no clear correlation suggesting that larger models perform better, which indicates that the issue isn't simply one of capacity or knowledge. Rather, it seems to stem from a systemic information bottleneck and a mismatch between training objectives. LVLMs are typically trained or fine-tuned to generate descriptive language or answer general questions, but they are not explicitly designed to retain all visual information. This training bias may lead them to prioritize what humans typically highlight when describing images, resulting in the unintentional filtering out of global structures and treating them as ``unimportant details''. As a result, these models have learned to ``see'' more like a describer than a true visual reasoner, focusing on local or salient objects rather than understanding the image in its entirety.

\section{Conclusion}
\label{conclusion}

In this paper, we have introduced TopoPerception, a diagnostic benchmark that rigorously and without shortcuts evaluates the global visual perception of LVLMs across different perceptual granularities. By focusing on the topological properties of images---global features that are independent of local characteristics---we have discovered that current LVLMs suffer from severe deficiencies in global visual perception. Our experiments on SOTA models show that even at the simplest evaluation levels, their performance is close to random chance, with answers based almost entirely on their intrinsic biases. This reveals a fundamental limitation masked by their fluent linguistic output on other problems.

These findings have profound implications for the future development of LVLMs. First, they highlight the urgent need to reconsider the architecture of multimodal models: simply ``stitching'' a fixed visual encoder to an LLM (with a minimal interface) may be insufficient for tasks requiring deeper image understanding. More expressive or iterative visual encoders, or mechanisms that allow the model to re-examine the image during reasoning, may be necessary. Second, the observation that larger scale and stronger reasoning capabilities can degrade performance suggests that multimodal models require careful calibration between reasoning and perception. When a model ``thinks'' in language, it may distort what it sees---a mechanism for ``fact-checking'' its reasoning against the visual input at each step could be beneficial.

In summary, as we strive to build AI that can genuinely ``perceive'' the world, TopoPerception provides a new lens for analyzing what our models see and what they miss. We hope it will serve as a diagnostic tool to guide the development of the next generation of vision-language models, leading to systems that are not only capable of fluently describing an image but also of grasping its deepest structural truths.

\section*{Acknowledgement}

This work was supported by the STI 2030--Major Projects 2021ZD0200300.

{
    \small
    \bibliographystyle{ieeenat_fullname}
    \bibliography{main}
}

\end{document}